%% file: main.tex
\documentclass{article}

\PassOptionsToPackage{numbers,compress}{natbib}

\usepackage[preprint]{neurips_2026}

\usepackage[utf8]{inputenc}
\usepackage[T1]{fontenc}
\usepackage[colorlinks=true, citecolor=newblue, linkcolor=newblue, urlcolor=black]{hyperref}
\usepackage{url}
\usepackage{booktabs}
\usepackage{amsfonts}
\usepackage{amsmath}
\usepackage{amssymb}
\usepackage{nicefrac}
\usepackage{microtype}
\usepackage{xcolor}
\usepackage{graphicx}
\usepackage[table]{xcolor}
\usepackage{multirow}
\definecolor{newblue}{rgb}{0.21,0.49,0.74}

\newcommand{\fyq}[1]{\textcolor{black}{#1}}

\newcommand{\best}[1]{\cellcolor{blue!25}\textbf{#1}}
\newcommand{\snd}[1]{\cellcolor{blue!10}#1}
\graphicspath{{figures/}}

\title{Afford-VLA: Action-Aligned Visual Planning via Internalized Affordance}

\author{
\textbf{Runze Wang}$^{1,*}$ \quad 
\textbf{Yuqian Fu}$^{2,}$\thanks{These authors have equal contributions.} \quad 
\textbf{Yu Li}$^1$ \quad 
\textbf{Tao Lin}$^3$ \quad 
\textbf{Tianwen Qian}$^4$ \\ 
\textbf{Mohamed Elhoseiny}$^2$ \quad 
\textbf{Bo Zhao}$^3$ \quad 
\textbf{Yanwei Fu}$^1$ \quad 
\textbf{Yu-Gang Jiang}$^1$ \quad 
\textbf{Xiangyang Xue}$^{1}$\thanks{Corresponding author.} \\
$^1$Fudan University, \quad $^2$KAUST, \quad $^3$SJTU, \quad$^4$East China Normal University \\
{\tt\small wangrz24@m.fudan.edu.cn},\quad 
{\tt\small yuqian.fu@kaust.edu.sa},\quad 
{\tt\small xyxue@fudan.edu.cn}
}

\makeatletter
\renewcommand{\@noticestring}{}
\makeatother

\begin{document}

\maketitle
\pagestyle{empty} 

\begin{abstract}
\fyq{
Vision-language-action (VLA) models have shown strong potential for generalist robot manipulation, yet they remain limited by insufficient spatial reasoning, particularly in determining \emph{where to interact} in complex visual scenes. While recent efforts introduce various forms of visual planning to address this issue, existing approaches either rely on global geometric cues, symbolic intermediate representations, or externally generated visual signals, which are often weakly coupled with downstream action prediction. In this work, we revisit visual planning in VLA systems and argue that effective planning should be \emph{local}, \emph{visually grounded}, \emph{internally generated}, and \emph{directly aligned with action}. Based on this insight, we propose \textbf{Afford-VLA}, a unified framework that internalizes task-conditioned affordance as an explicit visual planning interface within VLA models. Concretely, we introduce learnable \texttt{<AFF>} tokens to query task-relevant interaction regions, decode affordance masks from multimodal features, and convert them into compact embeddings that directly condition action generation. This design enables affordance to be both generated and utilized within the VLA, forming a tightly coupled perception–action pathway. To further support this integration, we adopt a training strategy that allows the affordance pathway to be jointly optimized with action prediction, improving its effectiveness for downstream control. We evaluate our method on multiple simulation benchmarks, including LIBERO, LIBERO-Plus, and SimplerEnv, achieving consistent state-of-the-art performance, along with strong real-world results. These findings demonstrate that internalizing affordance as action-aligned visual planning provides a powerful paradigm for improving VLA systems. Codes and Models will be released at \href{https://github.com/RZkiller/AffordVLA}{\textcolor{pink}{Afford-VLA}}.
}

\end{abstract}

\input{sections/intro}

\input{sections/related}

\input{sections/method}

\input{sections/experiments}

\section{Conclusion}
\fyq{
In this work, we revisit visual planning in vision-language-action (VLA) systems and highlight the importance of explicitly modeling \emph{where to interact} for robust manipulation. We identify four key properties of effective visual planning—locality, visual grounding, internal generation, and action alignment—and show that existing approaches fall short in jointly satisfying them.  To address this, we propose \textbf{Afford-VLA}, a unified framework that internalizes task-conditioned affordance as an explicit visual planning interface within VLA models. By enabling affordance to be both generated and utilized as part of the decision process, our approach establishes a tightly coupled perception–action pathway, allowing interaction regions to directly influence action generation. Extensive experiments on both simulation benchmarks and real-world settings demonstrate the effectiveness of our method. More broadly, our results suggest that internalizing affordance as action-aligned visual planning provides a promising direction for improving spatial reasoning and control in embodied AI systems.
}

\bibliographystyle{plainnat}
\bibliography{references}

\newpage
\appendix
\input{sections/appendix}


\end{document}

%% file: sections/intro.tex
\section{Introduction}
\label{sec:intro}
\fyq{Vision-language-action (VLA) models~\cite{brohan2022rt,zitkovich2023rt,o2024open,pertsch2025fast,kim2025fine,black2024pi_0,bjorck2025gr00t, wang2026oflow} have recently shown strong promise for generalist robot manipulation by mapping visual observations and language instructions to actions. However, existing VLA systems still share a fundamental challenge: spatial understanding, i.e., how actions relate to the visual scene. This challenge is particularly critical for manipulation, where successful execution depends on accurately reasoning about spatial relationships in complex, real-world environments.}

\begin{figure}[t!]
    \centering
    \includegraphics[width=1.\linewidth]{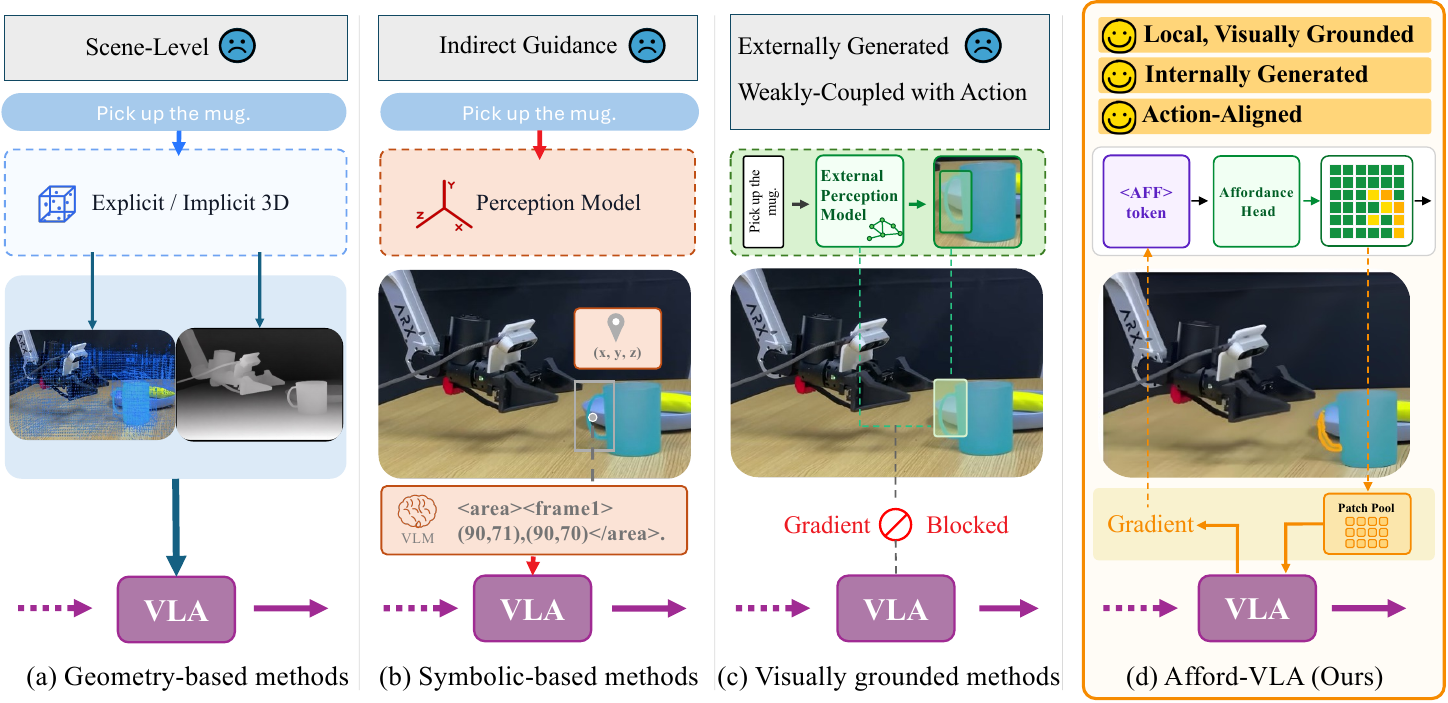}
    \vspace{-0.15in}
\caption{\fyq{\textbf{Comparisons of visual planning paradigms for VLA systems.}
(a) Geometry-based methods usually capture global scene-level cues.
(b) Symbolic-based methods use textual or token-based abstractions.
(c) Visually grounded methods capture explicit interaction regions, but are often externally generated or weakly coupled with action prediction.
(d) Afford-VLA learns local, visually grounded, internalized, and action-aligned affordance for direct action guidance. }}
\vspace{-0.2in}
\label{fig:teaser}
\end{figure}

\fyq{Many efforts have been made to improve spatial reasoning in VLA systems from the perspective of \emph{visual planning}, which we refer to as the ability to guide models to determine \emph{where to interact in the visual space}. As illustrated in Fig.~\ref{fig:teaser}(a-c), existing visual planning approaches can be roughly grouped into several representative types. i) Geometry-based methods~\citep{li2026pointvla,yuan2025depthvla,zhang20254d} leverage explicit or implicit 3D cues to enhance spatial awareness, but typically provide global, scene-level context rather than precise, task-conditioned interaction regions. 
ii) {Symbolic-based methods}~\citep{yuan2024robopoint,sundaresan2023kite,liu2024moka} provide more focused spatial guidance by grounding visual observations into intermediate symbolic representations, such as language descriptions and structured tokens, thus introducing an indirect guidance.
iii) Visually grounded methods~\citep{zeng2021transporter,zhao2025cot,huang2025roboground, li2025coa, nasiriany2025rt, wu2025afforddp,ma2025glover++} capture task-relevant interaction regions more explicitly through sparse or dense visual cues, such as masks. However, existing methods often rely on external perception modules or treat key-region localization as an independently supervised objective, which weakens its coupling with action learning and limits its direct integration into action prediction. 
These limitations suggest that how to achieve precise and actionable visual planning for VLAs remains an open challenge.}

\fyq{
We argue that effective visual planning for VLA systems should satisfy four key properties. 
(1) It should be \emph{local}, enabling VLAs to focus on task-relevant interaction regions.
(2) It should be \emph{visually grounded}, directly tied to visual evidence rather than mediated through abstract representations. 
(3) It should be \emph{internally learned} within the VLA model, rather than produced by external modules. 
(4) It should be \emph{action-aligned}, so that it can be directly consumed by the action model to influence downstream decision making. 
Together, these properties define a form of visual planning that bridges perception and action, providing a principled foundation for learning where to interact in VLAs.
}

\fyq{
To achieve these goals, we begin by considering the choice of visual planning representation. Among possible candidates, affordance naturally captures task-relevant interaction regions in a local and visually grounded manner, making it well suited for guiding manipulation.
The core idea of our approach is to treat task-conditioned affordance as an internal interface that bridges perception and action. Specifically, as illustrated in Fig. 1(d), we introduce learnable \texttt{<AFF>} tokens as part of the input to the VLA model, alongside image patches and textual instructions. Through the model’s attention mechanism, these tokens 
aggregate visual and language information and are used to decode task-conditioned affordance masks via a dedicated affordance head. 
To make affordance directly useful for action control, we further convert the predicted masks into compact embeddings through mask pooling over visual features. These embeddings are then fed into the action head, enabling interaction regions to directly influence action prediction. These designs allow affordance to be both generated and utilized within the VLA model as part of the decision process, resulting in a tightly coupled perception–action pathway. This distinguishes our approach from prior affordance-based methods~\citep{li2025coa, nasiriany2025rt, wu2025afforddp,ma2025glover++}, where affordance is typically produced externally or used only as an auxiliary signal. 
To further strengthen this coupling, we adopt a straight-through gradient estimator, allowing gradients from action prediction to propagate through the affordance pathway. 
This enables the affordance representation to be jointly optimized for both action prediction and dense supervision signals.
Based on these designs, we introduce \textbf{Afford-VLA}, a unified framework that internalizes task-conditioned affordance as an explicit visual planning interface, enabling \emph{local}, \emph{visually grounded}, \emph{internally generated}, and \emph{action-aligned} interaction modeling within VLA systems.
}

\fyq{
We conduct extensive experiments on multiple simulation benchmarks, including LIBERO~\cite{liu2023libero}, LIBERO-Plus~\cite{fei2025libero}, and SimplerEnv~\cite{li2024evaluating}, achieving state-of-the-art performance across all benchmarks (97.4\%, 78.1\%, and 58.1\%, respectively). Real-world manipulation experiments and ablation studies further validate the effectiveness of Afford-VLA and its core design.
}

\fyq{
Our contributions are summarized as follows:
\begin{itemize}
    \item We identify accurate \textbf{visual planning} as a key missing component in current VLA systems and formulate what constitutes effective visual planning through four properties: locality, visual grounding, internal generation, and action alignment.   
    \item We propose \textbf{Afford-VLA}, a unified framework that internalizes task-conditioned affordance as an explicit visual planning interface, enabling interaction regions to be directly learned and leveraged for action generation.
    \item We introduce an effective \textbf{training and integration} strategy that tightly couples affordance prediction with action learning, enabling the affordance pathway to be jointly optimized with downstream control objectives. Extensive experiments across multiple simulation benchmarks and real-world settings validate the effectiveness of our approach.
\end{itemize}
}

%% file: sections/related.tex
\section{Related Work}
\label{sec:related}

\noindent\textbf{Vision-Language-Action Models.}

Vision-language-action (VLA) models have emerged as a scalable paradigm for conditioning robot action generation on visual observations and language instructions. \fyq{Early works such as RT-1~\citep{brohan2022rt} and RT-2~\citep{zitkovich2023rt} demonstrated the potential of scaling transformer-based policies and integrating pretrained vision-language models into robotic control.} Subsequent efforts further expanded this paradigm through cross-embodiment data and open generalist policies, including Open X-Embodiment~\citep{o2024open}, Octo~\citep{team2024octo}, and OpenVLA~\citep{kim2024openvla}. More recent systems have diversified action decoding from vision-language representations, ranging from autoregressive action tokenization and improved tokenizers such as FAST~\citep{pertsch2025fast}, to parallel continuous regression as in OpenVLA-OFT~\cite{kim2025fine}, and flow-matching or diffusion-style action experts as in $\pi_0$~\citep{black2024pi_0} and GR00T~\citep{bjorck2025gr00t}.
\fyq{Despite these advances, most VLA systems still rely on the action decoder to implicitly infer task-relevant interaction regions from global representations, which can limit manipulation requiring precise spatial grounding. We therefore study how to equip VLAs with an explicit visual planning interface that provides task-conditioned guidance on \emph{where to interact}.}

\noindent\textbf{Visual Planning for Robot Control.}
\fyq{
Recent works have explored visual planning to improve spatial reasoning in robot control. Based on the spatial representations used to convey planning information, we broadly group existing methods into three categories. 
\textit{Geometry-based methods~\cite{li2026pointvla,yuan2025depthvla,zhang20254d}} anchor robot policies in the physical structure of the workspace through explicit or implicit 3D cues, such as point clouds, depth, and multi-view features. While these representations provide useful geometric awareness, they often operate at a global or scene-level scale and may not explicitly expose local, task-conditioned interaction regions. 
\textit{Symbolic-based methods~\cite{yuan2024robopoint,sundaresan2023kite,liu2024moka}} express spatial guidance through intermediate textual, symbolic, or token-based representations. Rather than directly grounding interaction cues in visual space, they represent them through symbolic abstractions such as language descriptions of object locations, resulting in an indirect mapping to action control.
}

\fyq{\textit{Visually grounded methods}~\citep{zeng2021transporter,zhao2025cot,huang2025roboground, li2025coa, nasiriany2025rt, wu2025afforddp,ma2025glover++} operate more directly in the visual space, using sparse cues such as points, bounding boxes, and trajectories, or dense cues such as masks and heatmaps. 
These visual cues can encode different forms of interaction semantics, among which \emph{affordance} is particularly relevant for manipulation, as it specifies which regions or motions support task-relevant actions.
Prior work has highlighted the promise of affordance from multiple perspectives: {CoA-VLA~\citep{li2025coa} shows that affordance-style reasoning can guide VLA action generation; RT-Affordance~\citep{nasiriany2025rt} treats affordance as a versatile intermediate representation for manipulation; AffordDP~\citep{wu2025afforddp} demonstrates its transferability for policy generalization; and GLOVER++~\citep{ma2025glover++} scales actionable affordance learning to large datasets.} 
However, existing affordance pipelines are often decoupled from downstream action learning: affordance is typically produced by an external perception module, or supervised as an independent intermediate target, rather than learned as an internal representation shaped by the action objective. 
As a result, the affordance pathway receives limited feedback from action prediction and is not always directly consumed by the action decoder.  In contrast, \textbf{Afford-VLA} represents affordance as an internal mask-based visual planning interface that is learned within the VLA model and directly conditions action generation.}

%% file: sections/method.tex
\begin{figure}[ht]
    \centering
    \vspace{-0.1in}
    \includegraphics[width=1.0\linewidth]{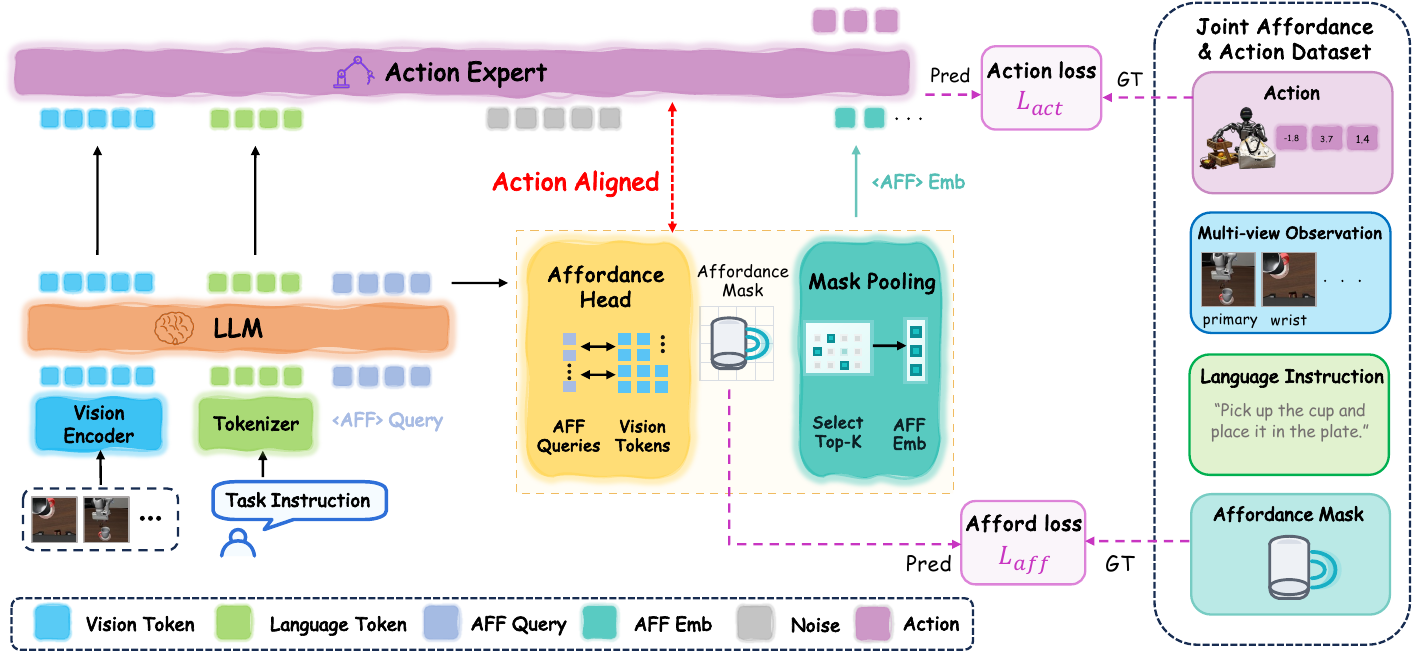}
\caption{
\textbf{Overview of Afford-VLA.}
\fyq{Afford-VLA formulates affordance as an internal visual planning interface for VLA systems. 
Learnable \texttt{<AFF>} query tokens first generate task-conditioned affordance masks from visual features, which are then converted into compact affordance embeddings through mask pooling to directly condition the action expert. 
The entire affordance pathway is jointly optimized with affordance loss and action prediction loss.}
}
    \label{fig:framework}
\end{figure}

\section{Methodology}
\label{sec:method}

\subsection{Formulating Affordance as Action-Aligned Visual Planning}
\label{sec:method:formulation}

We formulate the role of affordance in Afford-VLA from the perspective of \emph{action-aligned visual planning}. Rather than treating affordance as an external perception output or an auxiliary visualization target, we introduce it as an internal visual planning interface: it localizes task-relevant interaction regions and converts them into compact features that can be directly consumed by the action model.

\noindent\textbf{Formulation and Overview.} For clarity, we first describe Afford-VLA under a single-view observation, and later extend it to multi-view inputs in Sec.~\ref{sec:method:extensions}. As illustrated in Fig.~\ref{fig:framework}, given an RGB observation $I_t$, a language instruction $x$, and optionally a proprioceptive state $s_t$, a standard VLA model predicts a future action chunk $\mathbf{a}_{t:t+H}$ as:
\[
    p(\mathbf{a}_{t:t+H} \mid I_t, x, s_t),
\]
leaving the action head to implicitly infer task-relevant interaction regions from global vision-language representations. Afford-VLA instead introduces a task-conditioned visual focus variable $M_t \in [0,1]^{H_I \times W_I}$, which denotes the affordance region for the current observation and instruction. The policy is then conditioned on both the original inputs and the affordance:
\[
    p(\mathbf{a}_{t:t+H} \mid I_t, x, s_t, M_t).
\]
Concretely, the model first predicts affordance masks conditioned on visual observations and instructions, and then aggregates them into compact features that directly guide action prediction. In this way, $M_t$ is not treated as a standalone output, but as an internal intermediate representation that bridges spatial visual grounding and control.

The following sections describe each component in detail: Sec.~3.2 presents internal affordance mask generation, Sec.~3.3 introduces affordance-conditioned action prediction, and Sec.~3.4 details the action-aligned training strategy.

\subsection{Internal Affordance Mask Generation}
\label{sec:method:mask_gen}

Afford-VLA generates affordance masks internally rather than relying on an external perception module. 
Given an image observation $I_t$ and a language instruction $x$, the VLM first converts them into image tokens $Q_{\mathrm{img}}$ and language tokens $Q_{\mathrm{text}}$. 
We augment the multimodal sequence with a small set of learnable affordance query tokens, denoted as \texttt{<AFF>} tokens. 
Let $Q_{\mathrm{aff}} \in \mathbb{R}^{K_{\mathrm{aff}} \times C_{\mathrm{llm}}}$ denote these query tokens, where $K_{\mathrm{aff}}$ is the number of affordance queries and $C_{\mathrm{llm}}$ is the VLM hidden dimension. 
The augmented sequence is processed by the same VLM backbone:
\[
    [H_t, A_t]
    =
    f_{\mathrm{VLM}}
    \bigl(
        [Q_{\mathrm{img}}, Q_{\mathrm{text}}, Q_{\mathrm{aff}}]
    \bigr).
\]
Here, $H_t$ denotes the contextualized hidden states of the original image-language tokens, while $A_t \in \mathbb{R}^{K_{\mathrm{aff}} \times C_{\mathrm{llm}}}$ denotes the hidden states at the \texttt{<AFF>} token positions. 
Because the \texttt{<AFF>} tokens participate in the same self-attention layers as image and language tokens, their final states are conditioned on both the current visual observation and the instruction.

To ground these internal affordance states in image space, we decode them together with patch-aligned visual features. 
Let $P_t \in \mathbb{R}^{N \times C_{\mathrm{vis}}}$ denote the visual patch features extracted from the image encoder, where $N=H_pW_p$ is the number of image patches and $(H_p,W_p)$ is the patch grid size. 
An affordance head $\mathcal{D}_{\mathrm{aff}}$ takes the contextualized affordance states and dense patch features as input, and predicts patch-level affordance logits:
\[
    G_t
    =
    \mathcal{D}_{\mathrm{aff}}(A_t, P_t),
    \qquad
    G_t \in \mathbb{R}^{H_p \times W_p}.
\]
The affordance head is implemented as a lightweight query-patch grounding decoder. 
Intuitively, the \texttt{<AFF>} states encode what interaction evidence should be searched for under the current instruction, while the patch features preserve where such evidence appears in the image. 
The decoder couples these two sources and assigns an affordance logit to each image patch. 
We provide the detailed architecture of $\mathcal{D}_{\mathrm{aff}}$ in the supplementary material.

The resulting logits $G_t$ serve as the internal visual planning signal used by the subsequent action-conditioning module. 
During training, they are supervised against patch-level affordance labels; during action prediction, they are used to select action-relevant visual patches for mask pooling.

\subsection{Affordance-Conditioned Action Prediction}
\label{sec:method:action_prediction}

Given the affordance logits $G_t$ predicted by the affordance head, we convert them into an action-consumable representation through mask pooling. 
The goal is to expose the action head to visual features from the interaction-relevant region, rather than requiring it to infer such regions only from the full VLM hidden sequence.

We flatten $G_t$ into $g_t \in \mathbb{R}^{N}$ and select the top-$k$ patches with the highest affordance logits. 
This produces a binary selection mask $m_t \in \{0,1\}^{N}$:
\[
    m_{t,i}
    =
    \mathbb{I}
    \left[
        i \in \operatorname{TopK}(g_t, k)
    \right],
    \qquad i=1,\ldots,N .
\]

Using $m_{t,i}$, we aggregate the selected patch features $P_t$ and project them to the VLM hidden dimension:
\[
    r_t
    =
    W_{\mathrm{aff}}
    \left(
    \frac{1}{k}
    \sum_{i=1}^{N}
    m_{t,i} P_{t,i}
    \right),
    \qquad
    r_t \in \mathbb{R}^{C_{\mathrm{llm}}}.
\]

$W_{\mathrm{aff}}$ projects vision features to the VLM hidden dimension.
We refer to $r_t$ as the affordance embedding, which summarizes local visual evidence selected by the predicted affordance logits.
The affordance embedding $r_t$ is then appended to $H_t$ to form the \fyq{action-conditioning sequence} as $Z_t = [H_t; r_t]$ and predict the future action chunk with the action head:

\[
    \hat{\mathbf{a}}_{t:t+H}
    =
    f_{\mathrm{act}}(Z_t, s_t),
\]

In our implementation, $f_{\mathrm{act}}$ follows the flow-matching action head used in Isaac-GR00T~\citep{bjorck2025gr00t}, which predicts continuous action chunks by learning a conditional vector field over action trajectories.
We keep its architecture unchanged and only augment its conditioning sequence with the affordance embedding.
This design turns the affordance mask into an internal action-conditioning interface. 
Rather than treating the mask as a detached perception output, Afford-VLA uses it to select local visual evidence and injects the resulting affordance embedding directly into action generation.

\subsection{Action-Aligned Training}
\label{sec:method:training}

The previous sections describe how Afford-VLA predicts affordance logits and converts them into affordance embeddings for action prediction. 
We now describe how this pathway is trained to be action-aligned. 
The goal is not only to make the affordance head match dense mask supervision, but also to ensure that the predicted affordance regions are useful for downstream action generation.

\noindent\textbf{Action-to-affordance gradient pathway.}
In Sec.~\ref{sec:method:action_prediction}, mask pooling selects high-affordance patches from $G_t$ and summarizes their visual features into an affordance embedding $r_t$. 
However, hard top-$k$ selection is not differentiable with respect to the affordance logits. 
To allow the action objective to update the affordance head, we use a straight-through variant of mask pooling during joint training:
\[
    r_t = \Phi_{\mathrm{ST}}(P_t, G_t).
\]
In the forward pass, $\Phi_{\mathrm{ST}}$ is identical to the hard top-$k$ mask pooling defined in Sec.~\ref{sec:method:action_prediction}, producing an affordance embedding from a sparse set of selected patches. 
In the backward pass, the non-differentiable selection is replaced with a soft surrogate over patch logits, allowing the action loss to update $G_t$ and the affordance head. 
As a result, the predicted affordance regions are optimized not only for visual grounding, but also for downstream action prediction.

\noindent\textbf{Training objectives.}
Afford-VLA uses an affordance grounding loss and an action prediction loss. 
{Particularly, we construct a \textbf{Joint Affordance $\&$ Action Dataset} which provides the ground-truth affordance mask $Y_t$ for affordance supervision. Details of the dataset and mask construction are provided in the Appendix.}

The affordance grounding loss is defined as:
\[
    \mathcal{L}_{\mathrm{aff}}
    =
    \mathrm{BCEWithLogits}
    \left(
        G_t,
        Y_t
    \right).
\]

For action learning, the action head takes $Z_t$ and predicts the future action chunk:
\[
    \mathcal{L}_{\mathrm{act}}
    =
    \ell_{\mathrm{FM}}
    \left(
        f_{\mathrm{act}}(Z_t, s_t),
        \mathbf{a}_{t:t+H}
    \right),
\]
{where $\ell_{\mathrm{FM}}$ denotes the flow-matching training objective.}

\fyq{Through straight-through mask pooling, $\mathcal{L}_{\mathrm{act}}$ not only optimizes action prediction but also provides feedback to the affordance pathway.}

\noindent\textbf{Two-stage training.}
We use a two-stage training strategy for stability. 
In the first stage, we warm up the affordance head using only dense mask supervision $\mathcal{L}_{\mathrm{aff}}$.

This stage gives the affordance head a stable spatial grounding before its predictions are used to condition action generation.
In the second stage, mask pooling is performed using the affordance logits predicted by the model, rather than ground-truth affordance masks.
The affordance embedding is computed from the predicted logits $G_t$ using $\Phi_{\mathrm{ST}}$, and the action head is trained on the resulting affordance-conditioned sequence. 
The joint objective is $\mathcal{L}_{\mathrm{joint}} = \mathcal{L}_{\mathrm{act}} + \mathcal{L}_{\mathrm{aff}}$.

During this stage, ground-truth masks are used only for the affordance loss, not for constructing the affordance embedding. 
This reduces the train-inference mismatch, since the action head is trained with affordance embeddings generated from the model's own predicted logits, the same type of conditioning signal used at test time.

\noindent\textbf{Inference.}
At inference time, ground-truth affordance masks are not available. 
Afford-VLA predicts affordance logits $G_t$ from the current observation and instruction, applies hard top-$k$ mask pooling to obtain the affordance embedding $r_t$, and conditions the action head on $[H_t; r_t]$:
\[
    \hat{\mathbf{a}}_{t:t+H}
    =
    f_{\mathrm{act}}
    \left(
        [H_t; \Phi_{\mathrm{TopK}}(P_t, G_t)],
        s_t
    \right).
\]
The straight-through surrogate is used only during training. 
Inference uses the same sparse hard mask pooling as the forward path.

\subsection{Extensions and Implementation Details}
\label{sec:method:extensions}

\noindent\textbf{Multi-view extension.}
The single-view formulation naturally extends to multi-view robot observations. Given $\mathcal{I}_t=\{I_t^v\}_{v=1}^{V}$, we apply the same affordance generation and pooling process to each view independently. Each view produces an affordance mask $\hat{M}_t^v$ and an affordance embedding $r_t^v$. The action head is then conditioned on the original VLM hidden states together with all view-level affordance embeddings:
\[
    Z_t = [H_t; r_t^1; \ldots; r_t^V].
\]
This extension preserves the core Afford-VLA pipeline while allowing the model to integrate complementary cues from different camera views, such as wrist and external cameras.

%% file: sections/experiments.tex
\section{Experiments}
\label{sec:experiments}

\noindent\textbf{Hardware.}
Our experiments are conducted on NVIDIA H200 GPUs. Specifically, we train the LIBERO models with 4 H200 GPUs and the SimplerEnv models with 8 H200 GPUs, while all simulation inference and evaluation are performed with 4 H200 GPUs. For real-world experiments, we deploy the policy on a 6-DoF ARX X5 robot arm. The robot observes the scene through two Intel RealSense D435 cameras, consisting of a wrist-mounted camera and a primary third-person camera. Demonstrations for teleoperation are collected using a U-Arm~\citep{zou2025u} device.

\noindent\textbf{Implementation.}
We use Qwen3-VL-4B-Instruct~\citep{yang2025qwen3} as the VLM and extract pre-projector patch features from its native vision encoder for affordance decoding and mask pooling. The action head follows a GR00T-style flow-matching design with a DiT-B backbone, predicting 8-step 7-DoF action chunks with 4 inference steps. The affordance branch uses 4 learnable <AFF> queries per view and a lightweight two-way decoder with 2 layers, 8 heads, and hidden size 256. We train in two stages: first warming up the affordance branch with dense mask supervision while freezing the VLM and action head, then training the policy with predicted affordance embeddings using straight-through top-16 patch pooling, allowing action gradients to refine affordance prediction.

\noindent\textbf{Baselines.}
We compare with representative VLA policies, including OpenVLA~\citep{kim2024openvla}, OpenVLA-OFT~\citep{kim2025fine}, \(\pi_0\)~\citep{black2024pi_0}, \(\pi_0\)-FAST~\citep{pertsch2025fast}, and Isaac-GR00T~\citep{bjorck2025gr00t}. 
We also include recent visual-planning-oriented methods, such as CoA-VLA~\citep{li2025coa}, DepthVLA~\citep{yuan2025depthvla}, SpatialVLA~\citep{qu2025spatialvla}, and 4D-VLA~\citep{zhang20254d}.

\begin{table*}[t!]
\caption{%
  \textbf{Benchmark Evaluation.}
  Comparison of Afford-VLA with existing VLA models on the LIBERO and SimplerEnv benchmarks.  For SimplerEnv, ``Put Spoon", ``Put Carrot", ``Stack Block", and ``Put Eggplant" denote the tasks ``put spoon on towel", ``put carrot on plate", ``stack green block on yellow block", and ``put eggplant in yellow basket", respectively. Success rates (\%) are reported, with \best{best} and \snd{second-best} results highlighted.
}
\label{tab:benchmark_results}
\centering
\resizebox{\textwidth}{!}{%
\begin{tabular}{llccccc}
\toprule
\textbf{Benchmark} & \textbf{Method} & \multicolumn{5}{c}{\textbf{Success Rate (\%)}} \\
\cmidrule(lr){1-7}
\textbf{LIBERO} &  & \textbf{Spatial} & \textbf{Object} & \textbf{Goal} & \textbf{Long} & \textbf{Average} \\
\midrule
& Diffusion Policy~\citep{chi2025diffusion}                   & 78.5       & 87.5       & 73.5       & 64.8       & 76.1 \\
& OpenVLA~\citep{kim2024openvla}              & 84.7       & 88.4       & 79.2       & 53.7       & 76.5 \\
& $\pi_0$~\citep{black2024pi_0}               & \snd{96.8} & \snd{98.8} & \snd{95.8} & 85.2       & 94.2 \\
& $\pi_0$-FAST~\citep{pertsch2025fast}        & 96.4       & 96.8       & 88.6       & 60.2       & 85.5 \\
& CoT-VLA~\citep{zhao2025cot}                 & 87.5       & 91.6       & 87.6       & 69.0       & 83.9 \\
& OpenVLA-OFT~\citep{kim2025fine}             & 95.2       & 94.2       & 95.2       & \snd{93.2} & 94.5 \\
& GR00T-N1~\citep{bjorck2025gr00t}            & 94.4       & 97.6       & 93.0       & 90.6       & 93.9 \\
& GR00T-N1.5~\citep{bjorck2025gr00t}          & 96.5       & 98.5       & 91.0       & 91.5       & 94.4 \\
\cmidrule(lr){2-7}
& DepthVLA~\citep{yuan2025depthvla}           & 96.4       & 98.0       & \snd{95.8} & 89.2       & \snd{94.9} \\
& SpatialVLA~\citep{qu2025spatialvla}         & 88.2       & 89.9       & 78.6       & 55.5       & 78.1 \\
& 4D-VLA~\citep{zhang20254d}                             & 88.9       & 95.2       & 90.9       & 79.1       & 88.6 \\
& SP-VLA~\citep{li2025sp}                             & 84.4       & 85.6       & 75.4       & 54.2       & 74.9 \\
& VLA-OS~\citep{gao2025vla}                             & 87.0       & 96.5       & 92.7       & 66.0       & 85.6 \\
& CoA-VLA~\citep{li2025coa}                            & 85.3       & 93.1       & 85.8       & 55.0       & 79.8 \\
& \textbf{Afford-VLA (Ours)}                  & \best{97.8} & \best{99.6} & \best{97.6} & \best{94.6} & \best{97.4} \\
\midrule

\textbf{SimplerEnv} &  & \textbf{Put Spoon} & \textbf{Put Carrot} & \textbf{Stack Block} & \textbf{Put Eggplant} & \textbf{Average} \\
\midrule
& RT-1-X~\citep{o2024open}                         &  0.0       &  4.2        &  0.0        &  0.0        &  1.1 \\
& Octo-Base~\citep{team2024octo}                            & 15.8       & 12.5        &  0.0        & 41.7        & 17.5 \\
& OpenVLA-OFT~\citep{kim2025fine}                 & 34.2       & 30.0        & 30.0        & 72.5        & 41.8 \\
& RoboVLM~\citep{liu2025towards}                         & 50.0       & 37.5        &  0.0        & 83.3        & 42.7 \\
& Magma~\citep{yang2025magma}                           & 37.5       & 29.2        & 20.8        & 91.7        & 44.8 \\
& CogACT~\citep{li2024cogact}                           & \snd{71.7} & 50.8        & 15.0        & 67.5        & 51.3 \\
& SpatialVLA~\citep{qu2025spatialvla}                   & 20.8       & 20.8        & 25.0        & 70.8        & 34.4 \\
& TraceVLA~\citep{zheng2024tracevla}                    & 12.5       & 16.6        & 16.6        & 65.0        & 27.7 \\
& VideoVLA~\citep{shen2025videovla}                     & \best{75.0} & 20.8        & \best{45.8} & 70.8        & 53.1 \\
& $\pi_0$~\citep{black2024pi_0}                          & 29.2       & 62.5        & 29.2        & 91.6        & 53.1 \\
& $\pi_{0.5}$~\citep{intelligence2025pi_}                     & 49.3       & \snd{64.7}  & \snd{44.7}  & 69.7        & \snd{57.1} \\
& Isaac-GR00T-N1.6-Bridge~\citep{bjorck2025gr00t}     & 64.5       & \best{65.5} &  5.5        & 93.0        & \snd{57.1} \\
& \textbf{Afford-VLA (Ours)}                            & 66.6       & 54.2        & 14.6        & \best{96.8} & \best{58.1} \\
\bottomrule
\end{tabular}%
}
\end{table*}

\begin{table}[ht]
\caption{%
  \textbf{LIBERO-PLUS Benchmark Evaluation.}
  Comparison of our method with VLA models on the LIBERO-PLUS benchmark across seven perturbation
  categories: Camera, Robot, Language, Light, Background, Noise, and Layout.
  The results report average success rates (\%) across tasks.
  \best{Best} and \snd{second-best} results are highlighted.
}
\label{tab:main_results_plus}
\centering
\resizebox{0.95\textwidth}{!}{%
\begin{tabular}{lcccccccc}
\toprule
\textbf{Method} & \textbf{Camera} & \textbf{Robot} & \textbf{Language} & \textbf{Light} & \textbf{Background} & \textbf{Noise} & \textbf{Layout} & \textbf{Total} \\
\midrule
OpenVLA~\citep{kim2024openvla}              &  0.8        &  3.5        & 23.0        &  8.1        & 34.8        & 15.2        & 28.5        & 15.6 \\
OpenVLA-OFT~\citep{kim2025fine}       & \snd{56.4}  & 31.9        & 79.5        & 88.7        & 93.3        & 75.8        & \snd{74.2}  & \snd{69.6} \\
OpenVLA-OFT\_w~\citep{kim2025fine}    & 10.4        & 38.7        & 70.5        & 76.8        & \snd{93.6}  & 49.9        & 69.9        & 55.8 \\
OpenVLA-OFT\_m~\citep{kim2025fine}    & 55.6        & 21.7        & \snd{81.0}  & \snd{92.7}  & 91.0        & 78.6        & 68.7        & 67.9 \\
NORA~\citep{hung2025nora}                       &  2.2        & 37.0        & 65.1        & 45.7        & 58.6        & 12.8        & 62.1        & 39.0 \\
WorldVLA~\citep{cen2025worldvla}               &  0.1        & 27.9        & 41.6        & 43.7        & 17.1        & 10.9        & 38.0        & 25.0 \\
UniVLA~\citep{bu2025univla}                   &  1.8        & \snd{46.2}  & 69.6        & 69.0        & 81.0        & 21.2        & 31.9        & 42.9 \\
$\pi_0$~\citep{black2024pi_0}                & 13.8        &  6.0        & 58.8        & 85.0        & 81.4        & \snd{79.0}  & 68.9        & 53.6 \\
$\pi_0$-Fast~\citep{pertsch2025fast}        & \best{65.1} & 21.6        & 61.0        & 73.2        & 73.2        & 74.4        & 68.8        & 61.6 \\
RIPT-VLA~\citep{tan2025interactive}                & 55.2        & 31.2        & 77.6        & 88.4        & 91.6        & 73.5        & \snd{74.2}  & 68.4 \\
\midrule
\textbf{Afford-VLA (Ours)} & 56.0 & \best{56.8} & \best{91.5} & \best{96.8} & \best{97.0} & \best{80.9} & \best{78.9} & \best{78.1} \\
\bottomrule
\end{tabular}%
}
\end{table}

\subsection{Main Results}
\noindent\textbf{Simulation Benchmarks.} \fyq{We evaluate Afford-VLA on three simulation benchmarks. First, we use LIBERO~\cite{liu2023libero}, a standard language-conditioned manipulation benchmark, and report success rates on four suites: Spatial, Object, Goal, and Long, which respectively test spatial reasoning, object-level generalization, goal variation, and long-horizon execution. Second, we evaluate on SimplerEnv~\cite{li2024evaluating} under the WidowX/Bridge real-to-sim setting, which measures visual and spatial generalization in realistic tabletop manipulation scenarios. Finally, we assess zero-shot robustness on LIBERO-Plus~\cite{fei2025libero}, which extends LIBERO with controlled perturbations in object layout, camera viewpoint, robot initialization, language, lighting, texture, and sensor noise. Note that the model trained on LIBERO is directly evaluated on LIBERO-Plus without any fine-tuning.}

\fyq{
\noindent\textbf{Comparative Results.}
Tab.~\ref{tab:benchmark_results} reports the main simulation results on LIBERO and SimplerEnv. 
Afford-VLA achieves the best average performance on both benchmarks, setting new state-of-the-art  results with average success rates of 97.4\% on LIBERO and 58.1\% on SimplerEnv. On LIBERO, Afford-VLA consistently outperforms all prior methods across the four task suites, including strong general VLA baselines such as $\pi_0$, OpenVLA-OFT, and GR00T variants, as well as visual planning methods such as SpatialVLA and CoA-VLA. These results validate the effectiveness of our core idea: internalizing affordance within the VLA model as an explicit visual planning interface. }
\fyq{
Beyond the overall SOTA performance, we further observe that Afford-VLA performs particularly well on tasks requiring strong spatial grounding and interaction-region localization. For example, it achieves 97.8\% on Spatial, 99.6\% on Object, 97.6\% on Goal, and 96.8\% on Put Eggplant in Yellow Basket. This trend aligns with our design: by predicting task-conditioned affordance and converting it into action-conditioning features, Afford-VLA guides the policy to focus on where to interact, thereby improving manipulation performance. 
We also note that Afford-VLA remains weaker than some competitors on several SimplerEnv tasks, such as block stacking. This may be because such tasks rely more on trajectory dynamics and long-horizon coordination, where localized affordance guidance is less directly beneficial. Nevertheless, Afford-VLA achieves the best overall average performance.
}

\fyq{
\noindent\textbf{Zero-Shot Robustness on LIBERO-Plus.}
Tab.~\ref{tab:main_results_plus} reports zero-shot performance on LIBERO-Plus.  Afford-VLA achieves the best average performance, with a 78.1\% average success rate across seven perturbation types. The improvements are especially notable under perturbations that directly challenge visual planning. For example, on Layout, Afford-VLA reaches 78.9\%, showing that it can re-localize task-relevant objects and placement regions when the scene configuration changes. Afford-VLA also performs strongly under Background (97.0\%), Noise (80.9\%), and Light (96.8\%) perturbations. These variations can distract policies that condition action prediction on global visual features, whereas affordance-guided pooling compresses visual evidence around task-relevant interaction regions. As a result, Afford-VLA is less sensitive to irrelevant textures, sensor noise, and appearance shifts. 
These results demonstrate that internal, visually grounded affordance provides a robust perception--action interface under distribution shifts that alter visual appearance without changing the underlying manipulation intent.
}

\subsection{Main Ablation Studies}

\fyq{
To validate the roles of \emph{internalized} and \emph{action-aligned} affordance, we conduct an ablation study with four settings:
(a) a vanilla VLA baseline without affordance,
(b) externally generated affordance injected into the action head,
(c) internal affordance generation without affordance-conditioned action prediction, and
(d) our full Afford-VLA design with jointly learned affordance generation and action conditioning.
All variants are evaluated on LIBERO under the same training protocol.}

\fyq{
As shown in Tab.~\ref{tab:ablation_affordance_integration1}, comparing (a) and (b) shows externally generated affordance with direct action conditioning already improves the baseline, demonstrating the benefit of explicitly exposing interaction regions to the action model.
Comparing (a) and (c) further shows that internally learned affordance improves spatial grounding even without direct action conditioning, although the gain remains limited.
Most importantly, our full design (d) achieves the best performance by a clear margin.
Comparisons between (b) and (d) validate the importance of \emph{internalized affordance generation}, while comparisons between (c) and (d) highlight the benefit of \emph{action-aligned affordance integration}.
Together, these results validate the effectiveness of our design.
}

\begin{table}[t!]

\caption{%
\textbf{Ablation on internal affordance generation and action alignment.}
We compare different affordance integration strategies.
Results are reported on LIBERO using average success rate (\%).
}
\label{tab:ablation_affordance_integration1}
\centering
\resizebox{\textwidth}{!}{%
\begin{tabular}{l cc c}
\toprule
& \multicolumn{2}{c}{\textit{Design Properties}}
& \textit{Success Rate (\%)} \\
\cmidrule(lr){2-3}\cmidrule(lr){4-4}
\textbf{Integration Strategy}
& \textbf{Internalized Generation}
& \textbf{Action Alignment}
& \textbf{LIBERO} \\
\midrule
(a) Baseline (no affordance)                                  & $\times$   & $\times$   &   95.4   \\
(b) External Affordance w Action Condition       & $\times$   & \checkmark &        96.5  \\
(c) Internal Affordance w/o Action Condition       & \checkmark & $\times$  &  95.9  \\
\textbf{(d) Internal Affordance w Action Aligned (ours)} & \checkmark & \checkmark & \textbf{97.4} \\
\bottomrule
\end{tabular}%
}
\end{table}

\subsection{Real-World Experiments}

We further evaluate Afford-VLA on two real-world tabletop manipulation tasks, a placement task \emph{Cup-to-Plate} and a picking task \emph{Fork-in-Bowl}. 
 
We compare against strong VLA baselines, including OpenVLA-OFT and $\pi_0$ under the same setup, model initialization, and evaluation protocol. Task-relevant objects are randomized within a predefined workspace during testing. 
Each method is evaluated for 20 trials per task.

\begin{figure}[ht]
    \centering
    \vspace{-0.1in}
    \includegraphics[width=1.0\linewidth]{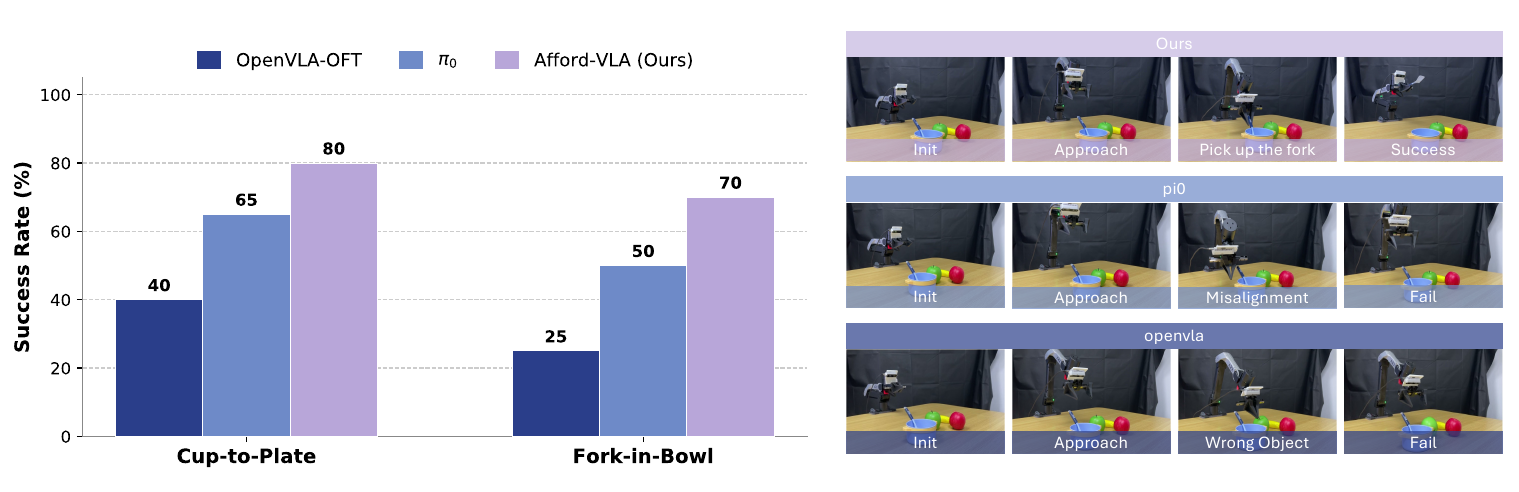}
    \vspace{-0.15in}
    \caption{\textbf{Real-world robot manipulation results.}
Left: quantitative comparison on two real-world manipulation tasks. 
Right: qualitative examples on the \textit{Fork-in-Bowl} task. }
    \vspace{-0.05in}
\label{fig:real_world_results}
\end{figure}

As shown in Fig.~\ref{fig:real_world_results}, Afford-VLA achieves the best performance on both tasks, reaching 80\% success rate on \emph{Cup-to-Plate} and 70\% on \emph{Fork-in-Bowl}. 
These results suggest that the proposed internal affordance planning improves real-world manipulation, especially in tasks requiring accurate localization of interaction regions.

{More experimental results, including more ablations, additional real-world experiments, and visualizations of the learned affordance masks, are provided in the Appendix.}

%% file: sections/appendix.tex
\section{Details of LIBERO-Plus Benchmark}

LIBERO-Plus is designed to evaluate whether vision-language-action policies remain reliable when the evaluation environment deviates from the clean conditions used in standard LIBERO evaluation. While the original LIBERO suites provide a useful testbed for language-conditioned manipulation, their standard evaluation protocol is largely based on fixed visual appearances, object layouts, camera configurations, and robot initial states. As a result, high success rates on the original benchmark do not necessarily indicate robustness to the distribution shifts commonly encountered in realistic manipulation scenarios.

To provide a more diagnostic evaluation, LIBERO-Plus systematically augments the original LIBERO task suites with controlled perturbations along multiple dimensions. The benchmark covers the four standard LIBERO suites, i.e., Spatial, Object, Goal, and Long, and expands them into a large-scale robustness testbed with 10,030 evaluation tasks. Each task preserves the underlying manipulation objective while modifying one aspect of the observation, scene configuration, language instruction, or robot state. This design allows us to measure whether a policy can maintain the same task intention under controlled changes, rather than relying on fixed visual or kinematic cues.

In our experiments, we use LIBERO-Plus as a zero-shot robustness benchmark. Specifically, models are trained on the original LIBERO training data and directly evaluated on LIBERO-Plus without any fine-tuning on the perturbed tasks. We report the average success rate under each perturbation category. This protocol is particularly relevant to Afford-VLA, since our method aims to improve action prediction by localizing task-conditioned interaction regions rather than depending only on global visual representations.

LIBERO-Plus contains seven categories of perturbations:

\begin{itemize}
    \item \textbf{Camera.}
    The camera setting is changed to test whether the policy can tolerate viewpoint shifts. Perturbations include moving the camera along the viewing direction, sampling camera poses around the scene, and changing the camera orientation while keeping the scene semantics unchanged. This category evaluates whether the model can re-ground the same manipulation target under different visual projections.

    \item \textbf{Robot.}
    The initial robot configuration is perturbed before task execution. Since the task instruction and scene content remain the same, successful execution requires the policy to adapt its action trajectory to a different starting state rather than memorizing a fixed motion pattern.

    \item \textbf{Language.}
    The natural-language instruction is rewritten while preserving the intended task. The rewrites may introduce irrelevant clauses, use alternative object or relation descriptions, or change the surface form of the instruction. This category tests whether the model is robust to linguistic variation and whether it grounds the instruction beyond memorized phrasing.

    \item \textbf{Light.}
    The illumination of the scene is modified by changing properties such as light intensity, direction, specular reflection, and shadows. These perturbations alter the visual appearance of objects and surfaces without changing the task structure, thereby testing robustness to low-level appearance shifts.

    \item \textbf{Background.}
    The visual appearance of the environment is changed by modifying scene textures or workspace surface textures, such as the tabletop. This category introduces visual distractors at the scene level and evaluates whether the policy can focus on task-relevant objects instead of overfitting to background appearance.

    \item \textbf{Noise.}
    Image-level sensor corruptions are applied to the observations, including blur-like and degradation-like effects. These perturbations mimic common camera artifacts and test whether the policy can still extract reliable task-relevant visual evidence from degraded inputs.

    \item \textbf{Layout.}
    The spatial arrangement of objects is changed by perturbing target object poses or adding irrelevant distractor objects. This category directly challenges spatial grounding and interaction-region localization, since the policy must identify where to act under a modified scene layout.
\end{itemize}

Among these perturbations, Camera and Layout are especially related to spatial grounding, as they require the policy to re-localize task-relevant regions under changed viewpoints or object configurations. Background, Light, and Noise mainly test robustness to visual appearance shifts. Robot perturbations evaluate sensitivity to initial kinematic states, while Language perturbations test instruction-level generalization. This decomposition enables a fine-grained analysis of how different VLA models fail under distribution shifts. In particular, the strong performance of Afford-VLA on Layout, Background, Noise, and Light perturbations suggests that internalized affordance provides a robust visual planning interface: by selecting task-relevant interaction regions and converting them into action-conditioning features, the policy becomes less sensitive to irrelevant visual changes and more capable of re-grounding its actions in perturbed scenes.

\section{Joint Affordance \& Action Dataset Construction}
We construct affordance mask supervision by augmenting the original LeRobot-format demonstrations with offline mask annotations. 
The original dataset is stored as episode-level trajectories, including parquet metadata and synchronized videos. 
We first convert each episode into frame-level samples, so that each observation frame can be independently processed by RAGNet, an affordance segmentation model. 
For each frame and camera view, RAGNet predicts an affordance mask, and the generated masks are saved in a separate directory.

We then merge the mask paths back into the original trajectory metadata. 
Specifically, for each frame, the corresponding mask paths are aligned with the original action, state, language instruction, and other metadata according to the episode and frame index. 
The resulting parquet files preserve the original LeRobot format but add view-specific mask path fields, e.g., for the primary and wrist camera views. 
During training, the native LeRobot dataloader directly loads the augmented parquet files and reads the corresponding mask images on demand. 
The loaded masks are used as dense affordance supervision for the affordance head, while the original action and state fields are unchanged.

\section{Implementation Details}
\label{sec:appendix_implementation}

We list the main hyperparameters of Afford-VLA in Tab.~\ref{tab:implementation_details}. We use Qwen3-VL-4B-Instruct as the vision-language backbone and a GR00T-style flow-matching action head to predict continuous robot actions. The action head follows a DiT-B configuration with 16 transformer layers and hidden dimension 1024. It predicts 7-DoF delta joint-position actions with a chunk length of 8, using the current proprioceptive state as additional input. During inference, we use 4 denoising steps for action generation.

For affordance modeling, we insert $K=4$ learnable \texttt{<AFF>} query tokens for each camera view. The queries are view-aware, implemented by adding a learnable view embedding to the shared affordance query embeddings. The hidden states at these \texttt{<AFF>} positions are decoded by a lightweight affordance head. The decoder uses two two-way attention layers with hidden dimension 256 and 8 attention heads, and predicts patch-level affordance logits on a $16 \times 16$ patch grid. We use dense pre-projector visual patch features from the Qwen3-VL vision encoder as the spatial features for mask decoding.

To condition action prediction on affordance, we use sparse Hard Top-$K$ patch pooling. Specifically, we select the top $k=16$ patches according to the predicted affordance logits in each view, average their visual features, and project the result to the VLM hidden dimension. The resulting affordance tokens are concatenated with the original VLM hidden states before being passed to the flow-matching action head. During training, this Top-$K$ operation is optimized with a straight-through estimator using a softmax surrogate with temperature $\tau=1.0$; during inference, we use pure hard Top-$K$ selection.

We train Afford-VLA in two stages. In the first stage, we warm up the affordance pathway for 4K steps while freezing the VLM backbone, action head, and region projection layer. In this stage, ground-truth affordance masks are used for region pooling. In the second stage, we initialize from the warmup checkpoint and train with predicted affordance masks using straight-through Top-$K$ pooling. The affordance head remains trainable in this stage, so the predicted affordance maps are optimized by $\mathcal{L}_{joint}$.

\begin{table}[t]
\centering
\caption{\textbf{Main implementation hyperparameters of Afford-VLA.}}
\label{tab:implementation_details}
\begin{tabular}{ll}
\toprule
Component & Setting \\
\midrule
VLM backbone & Qwen3-VL-4B-Instruct \\
Attention implementation & FlashAttention-2 \\
Action head & GR00T-style flow-matching DiT-B \\
Action / state dimension & 7 / 7 \\
Action representation & Delta joint position \\
Action chunk length & 8 \\
Inference denoising steps & 4 \\
Repeated diffusion steps & 4 \\
\midrule
AFF queries per view & 4 \\
Maximum number of views & 2 \\
Affordance decoder dimension & 256 \\
Affordance decoder layers / heads & 2 / 8 \\
Patch grid size & $16 \times 16$ \\
Pooling strategy & Hard Top-$K$ patch pooling \\
Top-$K$ patches & 16 \\
ST temperature $\tau$ & 1.0 \\
\midrule
Warmup steps & 4K \\
Second-stage steps & 140K \\
Optimizer & AdamW \\
Base learning rate & $2.5 \times 10^{-5}$ \\
VLM learning rate & $1.0 \times 10^{-5}$ \\
Action-head learning rate & $1.0 \times 10^{-4}$ \\
LR schedule & Cosine with min LR $1.0 \times 10^{-6}$ \\
Warmup steps for LR & 5K \\
Batch size per GPU & 16 \\
Image resolution & $224 \times 224$ \\
Training precision & Mixed precision \\
Distributed training & DeepSpeed ZeRO-2 \\
\bottomrule
\end{tabular}
\end{table}

\subsection{Compute Resources}
Afford-VLA is trained with mixed precision on NVIDIA H200 GPUs, each with 140GB memory. 
Table~\ref{tab:compute_resources} summarizes the compute resources used for training and evaluation. For real-world experiments, the policy is deployed for inference on a workstation with an NVIDIA RTX A6000 GPU with 48GB memory. 
Each real-world task is evaluated for 20 trials and takes approximately 1 hour, including policy inference and robot execution but excluding manual scene reset. 
All reported times are wall-clock times.

\begin{table}[h]
\centering
\caption{Compute resources used in our experiments.}
\label{tab:compute_resources}
\begin{tabular}{lcccc}
\toprule
Experiment & GPUs & Batch Size & Steps & Time \\
\midrule
LIBERO affordance warmup & 4 $\times$ H200 140GB & 16 & 4K & 0.5 h \\
LIBERO joint training & 4 $\times$ H200 140GB & 16 & 140K & 20 h \\
SimplerEnv affordance warmup & 8 $\times$ H200 140GB & 16 & 4K & 0.5 h \\
SimplerEnv joint training & 8 $\times$ H200 140GB & 16 & 200K & 48 h \\
LIBERO evaluation & 4 $\times$ H200 140GB & -- & -- & 2 h \\
LIBERO-Plus evaluation & 4 $\times$ H200 140GB & -- & -- & 48 h \\
SimplerEnv evaluation & 4 $\times$ H200 140GB & -- & -- & 4 h \\
\bottomrule
\end{tabular}
\end{table}

\section{Additional Experimental Details}

\subsection{Ablation on Mask Pooling Design}

\begin{table}[ht]
\centering
\caption{%
  \textbf{Ablation on mask pooling designs.}
  Hard Region Pooling preserves localized cues but blocks gradients to the predicted mask.
  Dense Soft Mask Pooling enables differentiability but dilutes affordance features.
  Sparse Top-$K$ ST Patch Pooling retains focused readout with end-to-end gradient propagation.
}
\label{tab:ablation_mask_pooling}
\small
\begin{tabular}{l cc c}
\toprule
& \multicolumn{2}{c}{\textit{Design Properties}}
& \textit{Success Rate (\%)} \\
\cmidrule(lr){2-3}\cmidrule(lr){4-4}
\textbf{Pooling Design}
& \textbf{Differentiable}
& \textbf{Localized Focus}
& \textbf{LIBERO} \\
\midrule
Hard Region Pooling              & $\times$   & \checkmark & 91.3 \\
Dense Soft Mask Pooling          & \checkmark & $\times$   &  96.0    \\
Sparse Top-$K$ ST Patch Pooling  & \checkmark & \checkmark & \textbf{97.4} \\
\bottomrule
\end{tabular}
\end{table}

To isolate the effect of pooling design from training strategy, all variants adopt the same two-stage warmup schedule; in stage 2, predicted masks are used for pooling across all designs.

As shown in Tab.~\ref{tab:ablation_mask_pooling}, the pooling mechanism plays a critical role in converting affordance masks into useful action-conditioning features. Hard Region Pooling achieves 91.3\% success rate. Although it preserves localized visual cues by selecting only the predicted affordance region, the hard mask operation is non-differentiable with respect to the predicted mask. As a result, gradients from the action loss cannot be propagated back to the affordance head through the pooling path. The affordance branch is therefore optimized mainly by mask supervision, which weakens its alignment with downstream action prediction.

Dense Soft Mask Pooling improves the success rate to 96.0\%, showing that differentiability is important for action-aligned affordance learning. Since the pooling weights are continuous, the action objective can update the predicted affordance logits through the pooled affordance embedding. However, this design averages features over a dense soft region, which inevitably mixes task-relevant interaction patches with surrounding background or distractor patches. The resulting affordance embedding becomes less focused, making it harder for the action head to receive sharp and localized interaction cues.

Sparse Top-$K$ ST Patch Pooling achieves the best performance, reaching 97.4\%. This design combines the advantages of the two alternatives. In the forward pass, it keeps a sparse hard Top-$K$ selection, so the action head receives focused features from the most affordance-relevant patches. In the backward pass, the straight-through estimator provides a soft surrogate gradient, allowing the action loss to update the affordance logits and the affordance head. Compared with Hard Region Pooling, it improves success rate by 6.1 points, confirming the importance of end-to-end action-to-affordance gradient flow. Compared with Dense Soft Mask Pooling, it further improves by 1.4 points, indicating that sparse localized readout is more effective than dense soft averaging for producing action-useful affordance embeddings. These results support our design choice: effective mask pooling should be both differentiable and spatially focused.

\subsection{Training Strategy Ablation}

\begin{table}[ht]
\centering
\caption{%
  \textbf{Ablation on training strategies for affordance conditioning.}
  We compare three strategies that differ in condition reliability, whether action gradients
  reach the Affordance Head, and train-inference alignment.
  Success rates (\%) on LIBERO.
}
\label{tab:ablation_training_strategy}
\small
\resizebox{\linewidth}{!}{%
\begin{tabular}{l ccc c}
\toprule
& \multicolumn{3}{c}{\textit{Design Properties}}
& \textit{Success Rate (\%)} \\
\cmidrule(lr){2-4}\cmidrule(lr){5-5}
\textbf{Training Strategy}
& \textbf{Cond.\ Reliability}
& \textbf{Action-to-Mask Feedback}
& \textbf{Train-Infer Alignment}
& \textbf{LIBERO} \\
\midrule
One-Stage w/ GT Pooling              & High   & $\times$   & $\times$   & 94.4 \\
One-Stage w/ Predicted Pooling       & Low    & \checkmark & \checkmark &   96.4   \\
\textbf{Two-Stage Warmup + Predicted Pooling (Ours)} & High & \checkmark & \checkmark & \textbf{97.4} \\
\bottomrule
\end{tabular}%
}
\end{table}
Here Mask Pooling Manner is fixed to the final design (Sparse Top-$K$ ST Patch Pooling) for all strategies, to isolate the effect of training strategy from pooling design. As shown in Table~\ref{tab:ablation_training_strategy}, the two-stage warmup strategy achieves the best performance by providing stable initial affordance conditioning and enabling end-to-end alignment between affordance perception and action learning.

\begin{table}[ht]
\caption{%
  \textbf{Ablation on affordance integration strategies.}
  Given affordance as the visual planning carrier, we compare five integration designs along three
  axes: whether affordance is \emph{explicit} (predicted as an intermediate output rather than only
  implicitly supervised), whether the affordance head is \emph{internal} to the VLA (jointly trained
  with the action backbone rather than a frozen external module), and whether the action-to-affordance
  pathway is \emph{differentiable} (allowing the action loss to update the affordance head).
  Success rates (\%) are reported on LIBERO.
}
\label{tab:ablation_affordance_integration}
\centering
\resizebox{\textwidth}{!}{%
\begin{tabular}{l ccc c}
\toprule
& \multicolumn{3}{c}{\textit{Design Properties}}
& \textit{Success Rate (\%)} \\
\cmidrule(lr){2-4}\cmidrule(lr){5-5}
\textbf{Integration Strategy}
& \textbf{Explicit Aff.}
& \textbf{Internal Head}
& \textbf{Differentiable}
& \textbf{LIBERO} \\
\midrule
(a) Baseline                                   & $\times$   & $\times$   & $\times$   &  95.4    \\
(b) Implicit Internal Affordance                          & $\times$   & \checkmark & $\times$   &  95.9    \\
(c) Explicit External Affordance                      & \checkmark & $\times$   & $\times$   &   96.5   \\
(d) Explicit Internal Affordance + Hard Pooling                & \checkmark & \checkmark & $\times$   &     91.3  \\
\textbf{(e) Explicit Internal Affordance} & \checkmark & \checkmark & \checkmark & \textbf{97.4} \\ 
\bottomrule
\end{tabular}%
}
\end{table}

\begin{figure}[ht]
    \centering
    \includegraphics[width=1.0\linewidth]{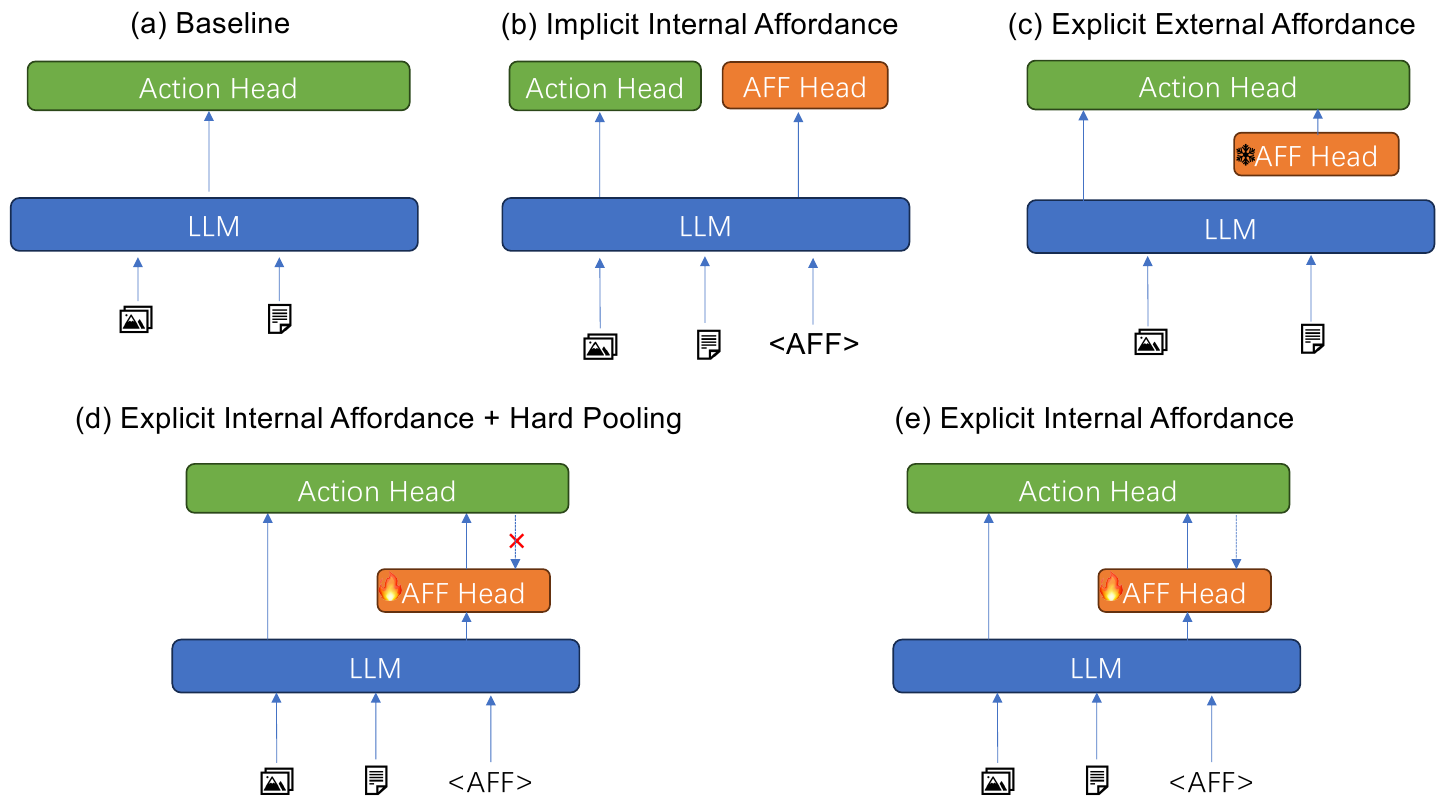}
  \caption{
\fyq{\textbf{Different affordance integration strategies for VLA systems.}
(a) Baseline VLA without introducing affordance modeling.
(b) Implicit internal affordance learns affordance within the VLA backbone but does not use it as condition of action prediction.
(c) Explicit external affordance uses an external affordance module to guide the action head.
(d) Explicit internal affordance with hard pooling directly injects affordance features into the action head, but blocks gradient propagation from action supervision.
(e) Our explicit internal affordance enables end-to-end action-aligned optimization, allowing action supervision to directly shape the affordance pathway.}
}
\label{fig:ablation_affordance_integration}
\end{figure}

Fig.~\ref{fig:ablation_affordance_integration} illustrates the compared integration strategies. 
Tab.~\ref{tab:ablation_affordance_integration} shows that affordance is useful for visual planning, but its benefit strongly depends on how it is integrated into the VLA. 
Implicit internal affordance brings only a small gain over the baseline, suggesting that auxiliary affordance supervision alone does not make the action head effectively use affordance. 
Explicit external affordance performs better, indicating that explicit localization cues are helpful; however, because the affordance module is external and not action-aligned, the improvement remains limited. 
Explicit internal affordance with hard pooling performs even worse than the baseline, since the non-differentiable pooling blocks action-loss gradients from updating the affordance head and turns the predicted mask into a brittle bottleneck. 
Our full design achieves the best performance by making affordance explicit, internal, and differentiably action-aligned, allowing affordance to serve as an effective visual planning representation.

\begin{figure}[t]
    \centering
    \includegraphics[width=0.92\linewidth]{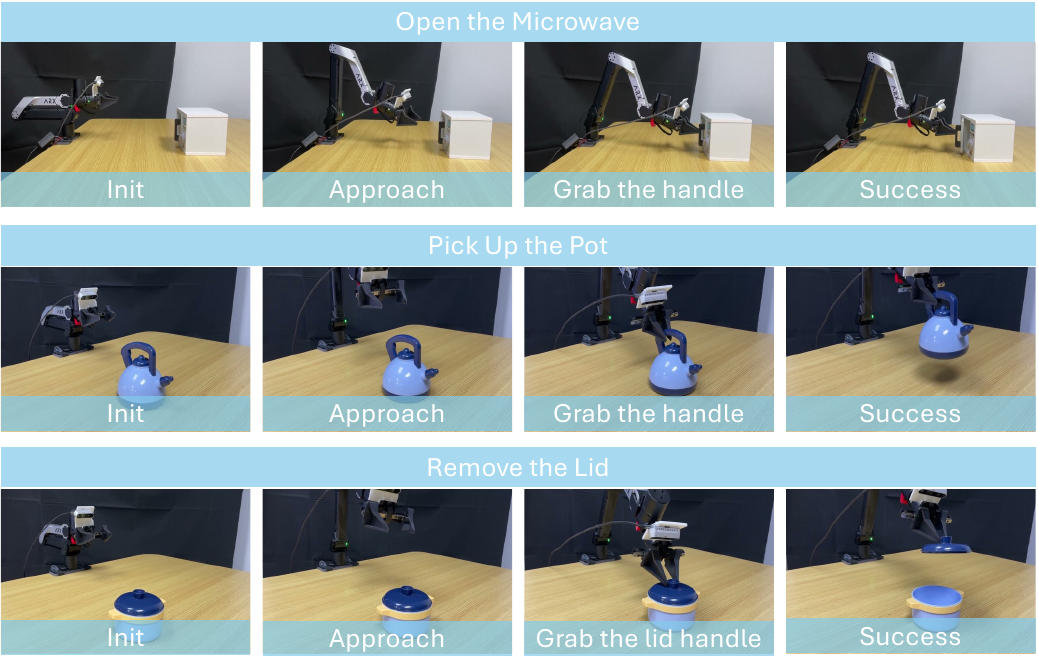}
    \caption{
    \textbf{Additional real-world robot manipulation results.}
    Qualitative rollouts of Afford-VLA on three real-world tasks:
    \emph{Open the Microwave}, \emph{Pick Up the Pot}, and \emph{Remove the Lid}.
    Each row shows a sequence from initialization to successful completion.
    }
    \label{fig:append_real_world_results}
\end{figure}

\begin{figure}[ht]
    \centering
    \includegraphics[width=1.0\linewidth]{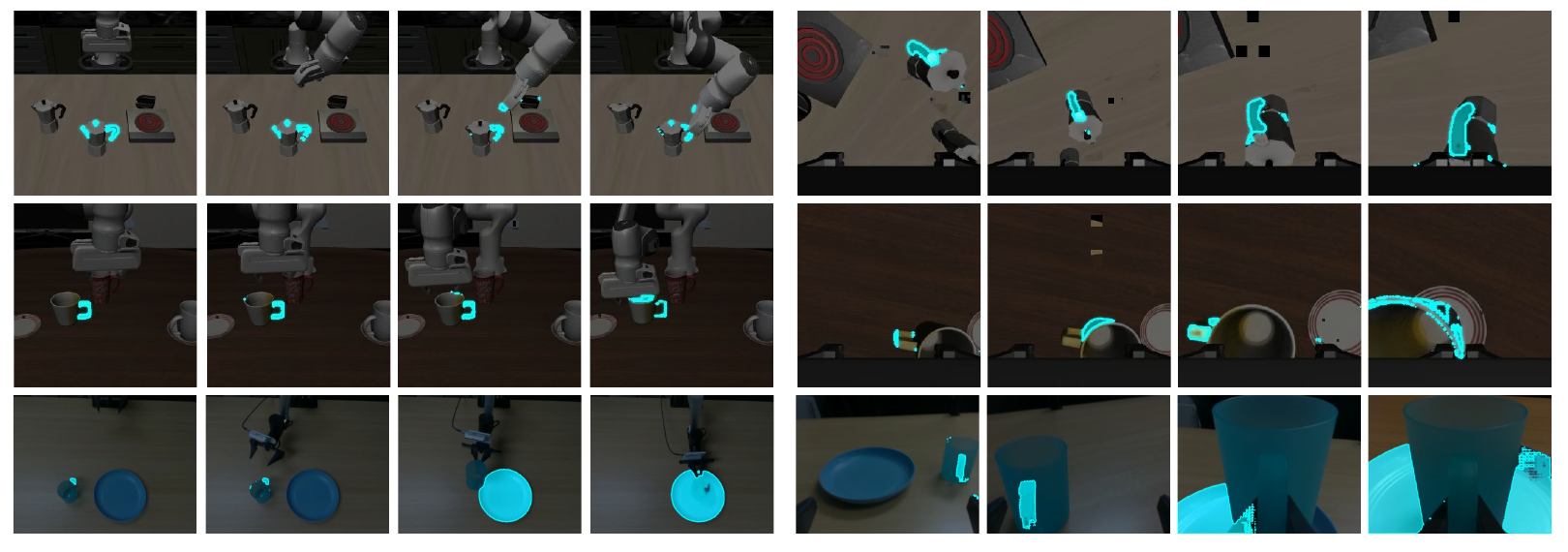}
\caption{
\fyq{\textbf{Visualization of learned affordance masks.}
Afford-VLA predicts task-conditioned affordance masks that dynamically focus on interaction-relevant regions during manipulation. 
}}
\label{fig:affordance_vis}
\end{figure}

\section{More Real-World Experiments}

We provide additional qualitative real-world results on three tabletop manipulation tasks: 
\emph{Open the Microwave}, \emph{Pick Up the Pot}, and \emph{Remove the Lid}. 
These tasks require the robot to localize small interaction regions, such as the microwave handle, pot handle, or lid handle, and maintain accurate spatial alignment throughout execution. 
As shown in Fig.~\ref{fig:append_real_world_results}, Afford-VLA is able to identify the task-relevant interaction regions and complete the manipulation sequence across different object configurations.

\section{More Analysis}
\subsection{Affordance Visualization}
Fig.~\ref{fig:affordance_vis} visualizes the affordance masks predicted by Afford-VLA in both simulated and real-world scenes. For each task, the left group shows the primary-view predictions and the right group shows the wrist-view predictions. Across different tasks, objects, and camera views, the predicted masks consistently concentrate on action-relevant interaction regions, such as graspable object parts, target containers, and contact areas. These qualitative results show that Afford-VLA learns strong task-conditioned visual grounding and can localize where the robot should act under diverse manipulation scenarios.

\subsection{Limitations}
\fyq{While Afford-VLA demonstrates strong performance across simulation benchmarks and real-world manipulation tasks, several limitations remain. First, our approach relies on affordance supervision during training, which in our current setup is derived from external model. While this provides strong spatial grounding signals without manual annotation, it introduces a dependency on the quality and bias of the underlying supervision source. As a result, the performance of Afford-VLA may be influenced;  Second, our formulation focuses on 2D visual affordance, leaving 3D affordance modeling unexplored. While incorporating 3D geometric information may further improve spatial reasoning and robustness, it also introduces non-trivial challenges, such as acquiring reliable 3D supervision, handling partial observations, and aligning 3D representations with action prediction in a unified framework. We leave these directions for future work, and view our current formulation as a first step toward validating the effectiveness of internalized affordance-based visual planning. Third, although we evaluate on multiple benchmarks and real-world tasks, the diversity of environments, object categories, and robot embodiments remains limited. As a result, further evaluation is needed to assess generalization to more complex, unstructured, or large-scale real-world scenarios. This limitation is common in current robot learning research due to the high cost and complexity of large-scale real-world data collection and evaluation.}

\subsection{Broader Impacts}
\fyq{This work aims to improve spatial reasoning and action grounding in vision-language-action (VLA) systems, which can benefit a range of applications in robotics. In particular, more reliable perception–action coupling may enhance robot manipulation in domains such as industrial automation, logistics, and assistive robotics, where accurate interaction with the environment is critical.
At the same time, deploying such systems in real-world environments introduces potential risks. Errors in affordance prediction or action generation may lead to unintended or unsafe behaviors, especially in unstructured or dynamic settings. These risks are particularly relevant when robots operate in close proximity to humans or handle delicate objects.
To mitigate these concerns, we emphasize that systems based on our approach should be deployed with appropriate safety constraints, monitoring mechanisms, and, where necessary, human oversight. We also note that our experiments are conducted in controlled simulation and limited real-world settings, and further validation is required before large-scale deployment. We hope this work contributes to the development of more reliable and interpretable robot learning systems, while encouraging careful consideration of safety and responsible use.}